%% file: main.tex
\documentclass[conference]{IEEEtran}
\usepackage{geometry}
\usepackage{booktabs} 
\usepackage{graphicx}
\usepackage{color}
\usepackage[linesnumbered,ruled]{algorithm2e}
\usepackage{ amssymb }
\usepackage{ mathrsfs }
\usepackage{float}
\usepackage{amsmath}
\usepackage{multirow}
\usepackage{transparent}
\usepackage{subcaption}
\usepackage{blindtext}

\makeatletter
\def\ps@IEEEtitlepagestyle{%
  \def\@oddfoot{\mycopyrightnotice}%
  \def\@evenfoot{}%
}
\def\mycopyrightnotice{%
  {\footnotesize The copyright belongs to me!\hfill}
  \gdef\mycopyrightnotice{}
}

\begin{document}
\date{}

\title{\Large\textbf{DRiLLS: Deep Reinforcement Learning for Logic Synthesis}}	

\makeatletter
\def\ps@IEEEtitlepagestyle{%
  \def\@oddfoot{\mycopyrightnotice}%
  \def\@evenfoot{}%
}
\makeatother
\def\mycopyrightnotice{%
  \begin{minipage}{\textwidth}
    \footnotesize
  \end{minipage}
  \gdef\mycopyrightnotice{}
}

\author{\normalsize
	\begin{tabular}[t]{c@{\extracolsep{1em}}c@{\extracolsep{1em}}c@{\extracolsep{1em}}c}
		\large Abdelrahman Hosny& \large Soheil Hashemi& \large Mohamed Shalan& \large Sherief Reda \\
		Computer Science Dept. & School of Engineering & Computer Science Dept. & School of Engineering \\
		Brown University  & Brown University & American University in Cairo  & Brown University \\
		Providence, RI & Providence, RI & Cairo, Egypt & Providence, RI\\
		abdelrahman\_hosny@brown.edu & soheil\_hashemi@brown.edu & mshalan@aucegypt.edu & sherief\_reda@brown.edu\\
\end{tabular}}
\maketitle

{\small\textbf{Abstract---
Logic synthesis requires extensive tuning of the synthesis optimization flow where the quality of results (QoR) depends on the sequence of optimizations used. Efficient design space exploration is challenging due to the exponential number of possible optimization permutations. Therefore, automating the optimization process is necessary. In this work, we propose a novel reinforcement learning-based methodology that navigates the optimization space without human intervention. We demonstrate the training of an Advantage Actor Critic (A2C) agent that seeks to minimize area subject to a timing constraint. Using the proposed methodology, designs can be optimized autonomously with no-humans in-loop. Evaluation on the comprehensive EPFL benchmark suite shows that the agent outperforms existing exploration methodologies and improves QoRs by an average of 13\%.
}}

\section{Introduction}
\label{sec:intro}
\input{1intro.tex}

\section{Design Space Exploration}
\label{sec:prev_work}
\input{2prev_work.tex}

\section{Background on Reinforcement Learning}
\label{sec:bg}
\input{3background.tex}

\section{Methodology}
\label{sec:meth}
\input{4methodology.tex}

\section{Experimental Results}
\label{sec:results}
\input{5results.tex}

\section{Conclusions}
\input{6conclusions.tex}
\label{sec:conc}

\section*{\sc Acknowledgments}
This work is supported by DARPA (HR0011-18-2-0032).

\bibliographystyle{IEEEtran}
\tiny
\bibliography{refbib}

\end{document}

%% file: 1intro.tex
Logic synthesis transforms a high-level description of a design into an optimized gate-level representation. Modern logic synthesis tools represent a given design as an And-Inverter Graph (AIG), which encodes representative characteristics for optimizing Boolean functions. Logic Synthesis mainly consists of three tightly-coupled steps, namely pre-mapping optimizations, technology mapping, and post-mapping optimizations. In the pre-mapping optimization phase, technology independent transformations are performed on the AIG to reduce the graph size resulting in a less total area, while adhering to a delay constraint. Next, in the technology mapping phase, the generic intermediate nodes are mapped to standard cells of a specific technology (e.g. ASIC standard cells). After that, post-mapping optimizations perform technology-dependent optimizations, such as up-sizing and downsizing.

Developing an efficient logic synthesis optimization flow proves to be an intricate task, requiring input from experienced designers. The complexity of designing such flows mainly arises from the exponentially large search space of the available transformations. In particular, with many transformation possibilities, different recurrence and permutations of such transformations can significantly affect the QoR~\cite{yu18, ziegler2016synthesis}. In addition, the increasing divergence and complexity in circuit designs have further complicated the design of optimization flows. It is crucial to note that there does not exist a pre-defined sequence of transformations that would generate best QoR for all possible circuits, and the optimization flows need elaborate tuning for each input.

At the same time, the advances in machine-learning (ML), and specifically reinforcement-learning (RL), have enabled autonomous agents to improve their capabilities in navigating a complex environment. Recently, successful implementations of such agents have shown to reach to human level or even outperform humans~\cite{silver2017mastering, jaderberg2018human}. For instance, AlphaGo was recently named the first computer to beat a professional human Go player~\cite{silver2017mastering}.

In this light, we propose a novel methodology based on RL that aims at producing logic synthesis optimization flows. Our contributions in this work are as follows:
\begin{itemize}
    \item We address the challenge of developing efficient design space exploration strategy. We map the problem of logic synthesis optimization to a game-like environment understandable by a reinforcement learning agent, and formulate a feature set extracted from the AIG characteristics. In addition, we derive a novel multi-objective reward function that aids the agent in minimizing the area subject to a delay constraint.
    \item We introduce DRiLLS (\textbf{D}eep \textbf{R}e\textbf{i}nforcemnet \textbf{L}earning-based \textbf{L}ogic \textbf{S}ynthesis), a novel framework based on reinforcement learning developed for generating logic synthesis optimization flows. Our methodology eliminates the need for a ``human expert'' tuning the synthesis parameters. It can be applied to any circuit design, without the need for a special setup.
    \item We demonstrate the capabilities of our proposed approach on the EPFL arithmetic benchmark suite \cite{epfl}. We compare our work against best results from the benchmark suite when mapped to a standard cell library, in addition to classical optimization algorithms such as greedy heuristics. Expert-developed flows are also evaluated for baseline comparison. We show that DRiLLS outperforms previous techniques \cite{epfl, expertscripts}.
\end{itemize}

The rest of the paper is organized as follows. First, in Section~\ref{sec:prev_work}, we define the problem and summarize relevant previous work. Next, in Section~\ref{sec:bg}, we present a background on RL that is utilized in our approach. Section~\ref{sec:meth} provides the motivation for our work as well as a detailed discussion on the proposed methodology. After that, we summarize our experimental results in Section~\ref{sec:results}. Finally, Section~\ref{sec:conc} summarizes the main contributions of this paper and provides possible future directions.

%% file: 2prev_work.tex
\subsection{Problem Statement}
\label{sec:problem}
In logic synthesis frameworks, there exist a rich set of primitive transformations, each optimizing the circuit using a different algorithm (e.g. balancing, restructuring). Permutations of  these optimizations  generate different QoR. Furthermore, different repetitions of the same  transformations affect the QoR and therefore result in an \textbf{exponentially growing search space}. Synthesis flows  for large circuits often have tens or hundreds of optimization commands.

We define $\mathbb{A} = \{a_1, a_2, ...a_n\}$ as the set of available optimizations in a logic synthesis tool. Let $k$ be the length of an optimization flow. Assuming that optimizations can be processed independently (e.g. no constraint for running $a_1$ before $a_2$), there exists $n^k$ possible flows. Yu \textit{et. al.} show that different flows indeed result in divergent area and delay results \cite{yu18}. While human experts have traditionally guided the search, the increasing complexity of the designs and synthesis optimizations have highlighted the need for an autonomous exploration methodology.

\subsection{Related Work}
Methodologies for design space exploration (DSE) of computing systems and EDA technology have received significant interest in the research community. On architectural level, Ipek {\it et al.} propose predictive models based on neural networks, to explore the design space of memory, processor, and multi-chip processor domains and predict the performance~\cite{Ipek2016}.
Similarly, Ozisikyilmaz {\it et al.} explore design space pruning by performance prediction of different computing configurations~\cite{Ozisikyilmaz}. In their work, they utilize three statistical models tuned on a small subset of the possible designs.
A learning-based methodology, relying on random forests for design space exploration of high-level synthesis flows is also proposed~\cite{Liu2013}.

More recently, Ziegler {\it et al.} proposed SynTunSys~\cite{ziegler2016synthesis}, a synthesis parameter tuning system which iteratively combines optimizations and focuses on the ``survivor set'' for further pursuit. Specifically, in each iteration, the candidates are assigned estimated costs and scenarios with the lowest cost values are evaluated. The cost estimator is then updated based on the learned costs~\cite{ziegler2016scalable}.
Taking a different approach, Yu {\it et al.} mapped the problem of logic synthesis design flow composition to a classification problem~\cite{yu18}. They then utilize convolutional neural networks to classify sample flows, encoded as pictures, to ``angel'' or ``devil'' flows. Therefore, for their work they require a fixed length for the optimization, and a large sample size of pre-defined optimization flows for training and tests. Our work is different from the previous work in that we propose to use a reinforcement learning agent to explore the search space for the purpose of optimizing particular synthesis metrics (e.g., area and delay), and therefore, enabling variable length optimization flows, without requiring sample flows for training. Next, we discuss relative background on RL that is used in this work.

In recent years, reinforcement learning (RL) agents have demonstrated immense capabilities in navigating complicated environments~\cite{mnih2015human, silver2017mastering}.
While earlier work using RL focused on domains with fully observable state space or where features could be handcrafted, Mnih {\it et al.} expanded these capabilities by introducing deep Q-networks (DQN)~\cite{mnih2015human}. Capitalizing on recent advances in deep neural networks, their agent achieves state-of-the-art performance in comparison to previous models, performing comparable to humans.
Further improving the capabilities of RL agents, in their work, Lillicrap {\it et al.} extended the action domain to continuous domain, targeting physical domains~\cite{lillicrap2015continuous}. 

%% file: 3background.tex
In this section, we briefly discuss the background necessary for developing our methodology. In reinforcement learning, an agent is trained to choose actions, in an iterative manner, that maximize its expected future reward. Formally,
\begin{itemize}
    \item At each iteration $k$, and based on the current state of the system $s_k$, the agent chooses an action $a_k$ from a finite set of possible actions $\mathbb{A}$.
    \item With the application of the action at step $k$, the system moves to the next state $s_{k+1}$ and a reward of $g(s_k,a_k)$ is then provided to the agent.
    \item The agent iteratively applies actions, changing the state of the system and getting rewards. It is then trained based on the collected experience to move toward maximizing its reward in future iterations. 
\end{itemize}

A policy is defined as a mapping $\mathcal{M}$ that, for each given state, assigns a probability mass function $\mathcal{M}(\cdot|a)$ for an action~\cite{konda2000actor}. There are two major categories for implementing the mapping $\mathcal{M}$: value-based and policy-based methodologies. In value-based methods (e.g. Q-learning) a value function is learned by the system that effectively maps $(state, action)$ pairs to a singular value~\cite{watkins1992q}, and picks the maximum over all possible actions. On the contrary, in policy-based methods (e.g. policy gradient), the optimization is performed directly on the policy ($\mathcal{M}$)~\cite{sutton2000policy}. Actor Critic algorithms~\cite{konda2000actor}, as a hybrid class, combine the benefits of both aforementioned classes.

In actor critic methods,  a tunable critic network provides a measure of how good the taken action is (similar to a reward function), while the tunable actor network chooses the actions based on the current state. More formally defined, the actor policy function is of the form $\pi_{\theta}(s,a$), and the critic function is of the form $\hat{q}_w(s,a)$; where $s$, and $a$ represent the state and the action, while $\theta$, and $w$ represent the tunable parameters within each network. Therefore, there exist two sets of parameters, one for each network, that need to be optimized. The gradient optimization for the critic network is performed as,

\begin{equation}
    \Delta w = \beta\delta\nabla_w \hat q_w(s_k,a_k)
\end{equation}
where $\beta$ sets different learning rate for policy and value. $\delta$ is the temporal difference error, which is defined as

\begin{equation}
    \delta = R(s,a) + \gamma \hat q_w(s_{k+1}, a_{k+1}) - \hat q_w(s_k,a_k)
\end{equation}

\noindent where $\gamma$ is the discount factor. Similarly, the gradient optimization for the policy update (actor network) is then defined as

\begin{equation}
\Delta \theta = \alpha \nabla_\theta (\log\pi_\theta(s,a))\hat q_w(s,a)
\end{equation}
where $\alpha$ sets the learning rate. Note that actor network policy update is a function of the critic network as well, which allows it to take into consideration not only the current state of the environment, but also the history of learning 
from the critic network.

While very effective, actor critic models can suffer from high variability in action probabilities. Advantage functions are proposed as a solution to reduce this variability. The advantage function is defined as

\begin{equation}
    A(s,a) = Q(s,a) - V(s)
\end{equation}
where $Q(s,a)$ represents the Q value for action $a$ in state $s$, and $V(s)$ represents the average value for the given state. In this work, we do not want to compute $Q(s, a)$. Instead, we formulate an estimate of the advantage function as

\begin{equation}
    \label{eq-discount-reward}
    A(s) = r+ \gamma V(s') - V(s)
\end{equation}

\noindent where $r$ is the current reward and $\gamma$ is the discount factor. This achieves the same result without learning the $Q$ function \cite{konda2003onactor}. Next, we describe the proposed DSE methodology based on reinforcement learning.

%% file: 4methodology.tex
\label{sec:malls}

\begin{figure}[t!]
		\includegraphics[scale=0.65]{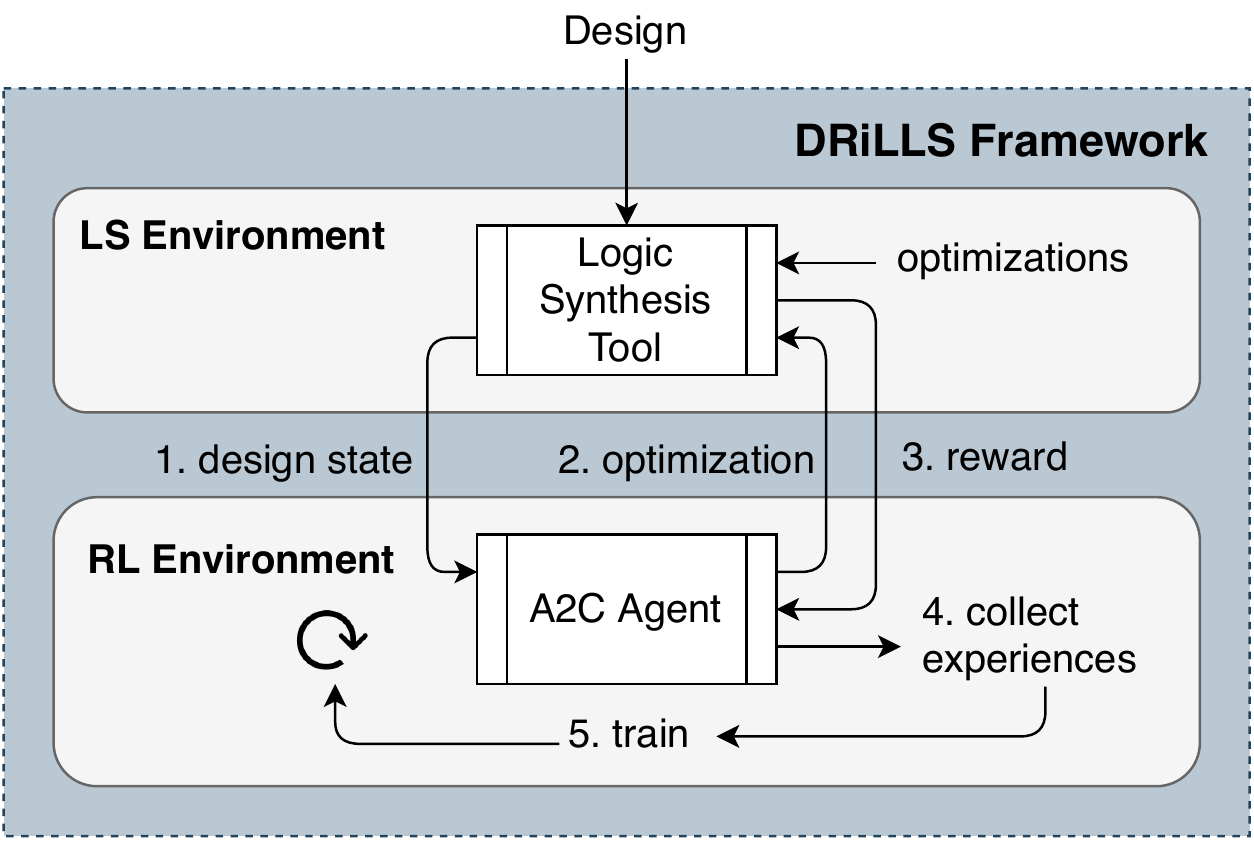}
		\centering
		\caption{The architecture of DRiLLS Framework. Numbers on the arrows represent the workflow of our methodology, and are illustrated separately in the subsections below.}
		\label{fig:architecture}
\end{figure}

DRiLLS, standing for \textbf{D}eep \textbf{R}e\textbf{i}nforcemnet \textbf{L}earning-based \textbf{L}ogic \textbf{S}ynthesis, effectively maps the design space exploration problem to a game environment. Unlike most reinforcement learning environments where gamification drives the behavior of the environment, the task here involves combinatorial optimization on a given circuit design. This makes it challenging to define the state of the game (i.e. environment) and the long-term incentives for an agent to explore the design space and not to fall into local minimums. 

Figure~\ref{fig:architecture} depicts the architecture of our proposed methodology. There are two major components in the framework: {\it Logic Synthesis} environment, which is a setup of the design space exploration problem as a reinforcement learning task, and {\it Reinforcement Learning} environment, which employs an {\it Advantage Actor Critic agent (A2C)} to navigate the environment searching for the best optimization at a given state. Next, we discuss both components and the interaction between them in details.

\textbf{1. Design State Representation.} In order to model combinatorial optimization for logic synthesis as a game, we define the \textit{state} of the logic synthesis environment as a set of metrics retrieved from the synthesis tool on a given circuit design and used as a feature set for the A2C agent. As previously discussed, the {\it state} also represents the reaction of the environment to an {\it optimization} suggested by the second component of our framework, namely the \textit{Agent}. Specifically, we extract the following state vector:

\small
\centerline{$\mathbf{AIG\ state} = \begin{bmatrix}\#\ primary\ I/O \\ \#\ nodes \\ \#\ edges \\ \#\ levels \\ \#\ latches \\ \%\ ANDs \\ \%\ NOTs \end{bmatrix}$.}

\vspace{0.1in}
\noindent To keep the states within a specific range, as required by the agent's neural networks, we normalize all state values by their corresponding values for the initial input design. Normalization is also a requirement for model generalization so that can be applied to unseen designs. On the one hand, optimizations change all elements in the \textit{state} vector, except for the number of primary inputs and outputs. On the other hand, values in the $state$ vector depict representative characteristics of the circuit. For example, a large $\#\ nodes$ value directs the agent towards reducing the number of nodes,  which is achieved by restructuring the current AIG and maximally sharing the other nodes available in the current network (e.g. $resub$ and $refactor$ commands in ABC). Moreover, a large $\#\ levels$ value steers the agent towards choosing a $balance$ transformation. Hence, the \textit{state} vector is representative of the circuit design at a given optimization step, and is aligned with the optimization space as we will discuss next.

\textbf{2. Optimization Space.} The agent explores the search space of seven primitive transformations, within ABC synthesis framework \cite{mishchenko2007abc}. Specifically, $\mathbf{\mathbb{A}}=\{$resub, resub -z, rewrite, rewrite -z, refactor, refactor -z, balance$\}$. The first six transformations target size reduction of the AIG, while the last one (balance) reduces the number of levels. These transformations manipulate the $state$ vector representation discussed above, and are appropriate for the reward function illustrated next.

\begin{table}[!t]
\footnotesize
\centering
\caption{Formulation of the multi-objective reward function. {\it \textbf{Decr.}} stands for Decrease and {\it \textbf{Incr.}} stands for Increase.}
\label{reward-table}
\begin{tabular}{|c|l|l|c|c|c|}
\hline
\multicolumn{3}{|l|}{\multirow{2}{*}{}} & \multicolumn{3}{c|}{\textbf{\begin{tabular}[c]{@{}c@{}}Optimizing\\ (Area)\end{tabular}}} \\ \cline{4-6} 
\multicolumn{3}{|l|}{} & \multicolumn{1}{l|}{\textit{Decr.}} & \multicolumn{1}{l|}{\textit{None}} & \multicolumn{1}{l|}{\textit{Incr.}} \\ \hline
\multirow{4}{*}{\textbf{\begin{tabular}[c]{@{}c@{}}Constraint\\ (Delay)\end{tabular}}} & \multicolumn{2}{l|}{Met} & +++ & \textit{0} & \textit{-} \\ \cline{2-6} 
 & \multirow{3}{*}{Not Met} & \textit{Decr.} & +++ & ++ & + \\ \cline{3-6} 
 &  & \textit{None} & ++ & 0 & -\ - \\ \cline{3-6} 
 &  & \textit{Incr.} & - & -\ - & -\ -\ - \\ \hline
\end{tabular}
\end{table}

\textbf{3. Reward Function.} We define a multi-objective reward function that takes into account the change in both design area and delay. In particular, the agent is rewarded for reducing the design area, while keeping the delay under a pre-specified constraint value. Table \ref{reward-table} shows the reward formulation of this function. For each metric (design area or delay), a transformation would decrease, increase or make no change to the metric. Accordingly, we give the highest reward (represented as +++) for a transformation performed on a given AIG state that reduces the area and meets the delay constraint. We give the lowest negative reward when the transformation performed increases the design area and delay, while not meeting the constraint. Between the two extremes, the values and magnitudes of the reward have been chosen carefully to aid in the agent exploration. Essentially, we prioritize meeting the delay constraint. When not met, a positive reward is also given if the delay improved (i.e. decreased). This reward strategy prevents the agent from receiving negative reward in all attempts in cases where the delay constraint was too tight for the design to meet. Moreover, when the area increases and the delay decreases (but not meeting the constraint), a small positive reward is given as the agent is trying to learn from not meeting the constraint. This reward formulation has proved to be efficient as we will discuss in the next section.

\begin{algorithm}[t]
    \small
    \SetKwBlock{Begin}{begin}{end}
    \SetKwInOut{Input}{Input}
    \SetKwInOut{Output}{Output}
    \Input{Design, Primitive\ Transformations}
    \Output{Optimization\_Flow}
    $env$ = Initialize(LS\_Env);\\
    $agent$ = Initialize(A2C);\\
    \For{$episode$ = $1$ to $N$}
    {
        $episode\_design\_states$ = [];\\
        $optimization\_sequence$ = [];\\
        $synth\_rewards$ = [];\\
        $design\_state$ = $env$.reset();\\
        \For{iteration = $1$ to $k$ }
        {
            $opt\_probs$ = agent.ActorForward($design\_state$);\\
            $primitive\_opt$ = RandomChoice($opt\_prob$);\\
            {[$next\_design\_state$, $synth\_reward$]} = $env$.perform($primitive\_opt$);\\
            $episode\_design\_states$.append($design\_state$);\\
            $optimization\_sequence$.append($primitive\_opt$);\\
            $synth\_rewards$.append($synth\_reward$);\\
            $design\_state$ = $next\_design\_state$;\\
        }
        $episode\_rewards$ = DiscountRewards($synth\_rewards$, $gamma$) \\
        $loss$ = agent.OptimizerForward($episode\_design\_states$, $optimization\_sequence$,  $episode\_rewards$);\\
        $agent$.update($loss$);\\
        log($episode$);\\
    }
    \caption{DRiLLS Framework}
    \label{alg:rl_method}
\end{algorithm}

\textbf{4. Collecting Experiences.} Algorithm~\ref{alg:rl_method} summarizes the operation of our proposed methodology. Here, lines 1 and 2 initiate the logic synthesis environment and the agent, respectively. Next, the agent is trained over the span of $N$ episodes, where in each episode the logic synthesis environment is restarted; i.e. the original input design is reloaded (line 7). Next, in lines 8-16, the agent iteratively suggests a sequence of $k$ primitive optimizations to produce the optimization flow. More specifically, first, in line 9, the agent computes the probability distribution of choosing one primitive optimization from the optimization space, $\mathbf{\mathbb{A}}$. 
Then, in line 10, one of the primitive optimizations is selected according to the probability distribution calculated in line 9.
Next, in line 11, the selected optimization is executed to determine its effect on the $design\_state$. In addition, the reward is computed using the reward function in Table \ref{reward-table}. After that, we store the synthesis state, the optimization performed and the reward in the pre-initialized variables. Finally, we transition the state of the agent to the state after performing the optimization. The number of iterations is capped by $k$ to provide the game with an elimination condition, and as the optimization improvements on a given circuit design fade out in later iterations. After all iterations are performed, we train the A2C agent from the collected experiences as we will discuss next. 

\textbf{5. A2C Agent Training.} The training step starts with discounting the delay rewards over iterations in order to give earlier iterations a higher priority in choosing a good optimization (line 17). After that, in lines 18-19 the loss is computed and the actor and critic networks are trained to minimize the loss value as described next. As discussed in Section \ref{sec:bg}, the agent has a hybrid policy-based and value-based networks, called actor and critic respectively. Both networks have an input layer of size equal to the {\it AIG state} vector length. In addition, a reward, $r$ is passed to the critic network for training, and a discounted reward is passed to the actor network (Equation \ref{eq-discount-reward}). The actor network outputs probability distribution over the available transformations. Therefore, the output layer in the actor network has a size equal to the size of $\mathbb{A}$. Since the agent is initialized with random parameters, transformations chosen in the start of the training process do not necessarily represent a good choice. Parameters of both networks are updated to reduce the loss using a gradient-based optimizer. This process is then repeated for a pre-defined number of times (called episodes), during which the agent is trained to predict improved optimization flows. In fact, the choice of a hybrid reinforcement learning architecture is suited for combinatorial optimization tasks as it gives the agent an opportunity to explore diverse optimization sequences, yet maintain a path towards optimal designs.

\captionsetup[sub]{font=footnotesize}
\begin{figure*}
    \begin{subfigure}[t]{0.33\textwidth}
        \includegraphics[scale=0.17]{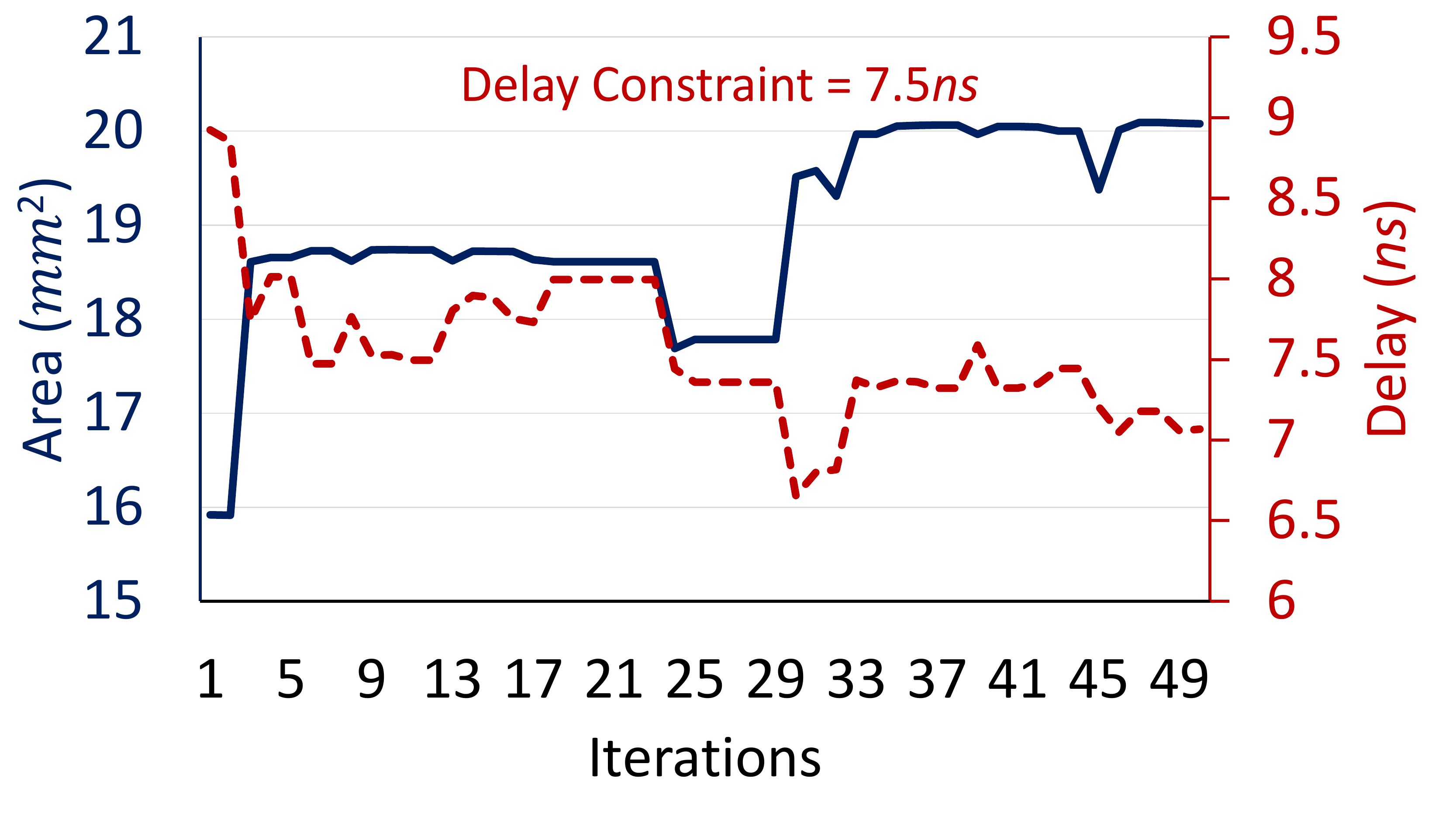}
        \caption{Log2}%
    \end{subfigure}
    \begin{subfigure}[t]{0.33\textwidth}
        \includegraphics[scale=0.17]{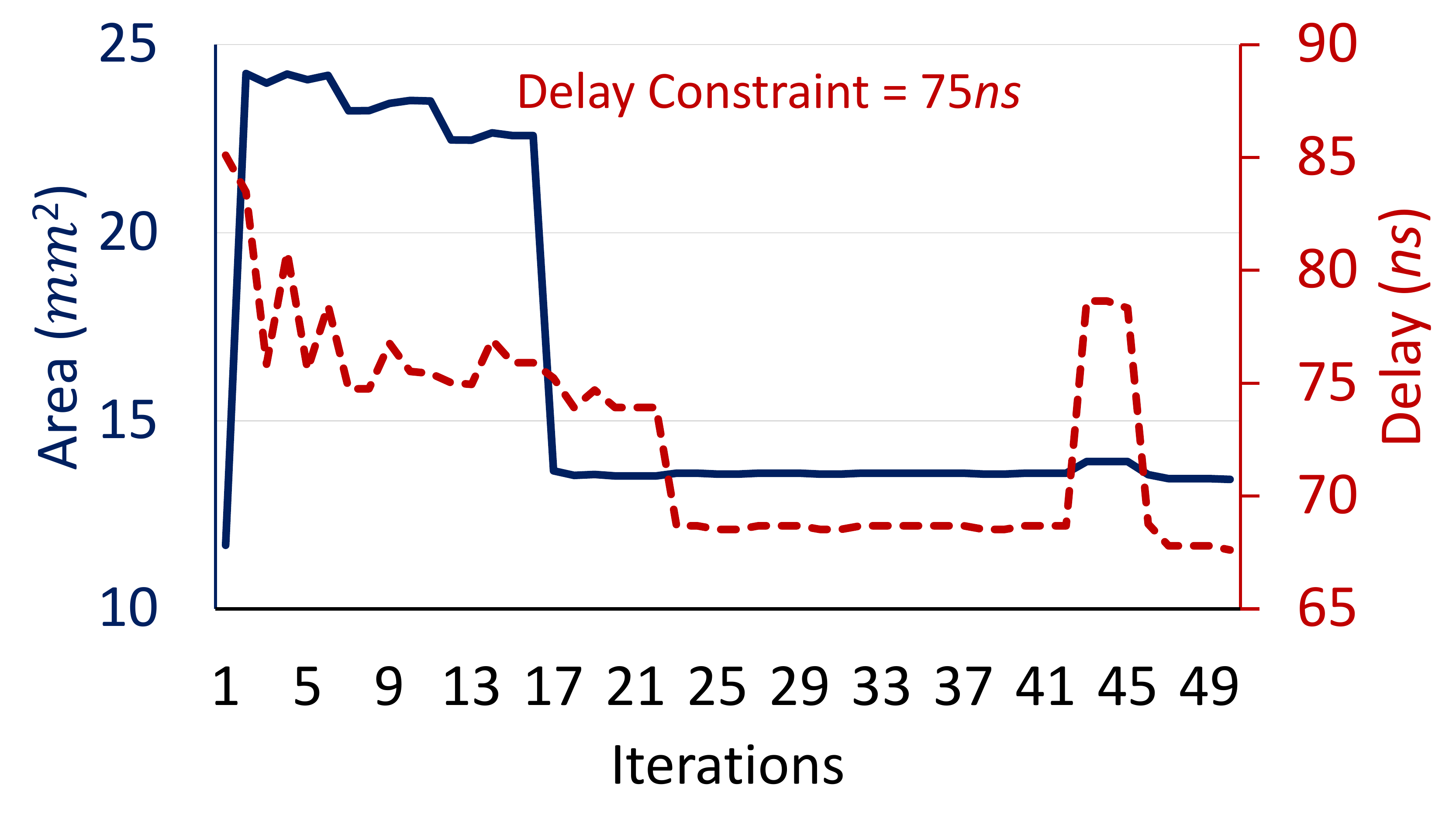}
        \caption{Divisor}
    \end{subfigure}
    \begin{subfigure}[t]{0.33\textwidth}
        \includegraphics[scale=0.17]{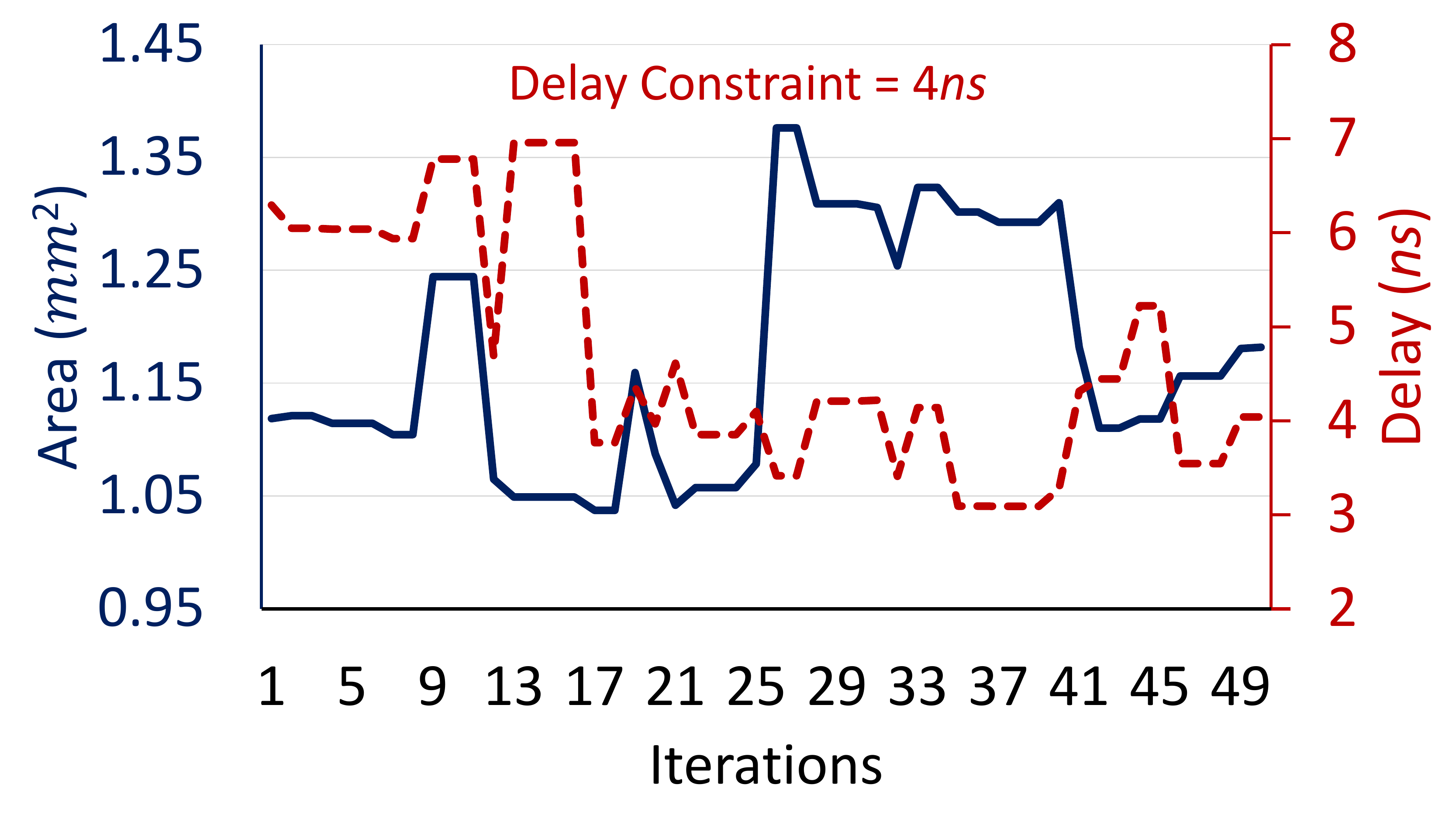}
        \caption{Max}
    \end{subfigure}
    \begin{subfigure}[t]{0.33\textwidth}
        \includegraphics[scale=0.17]{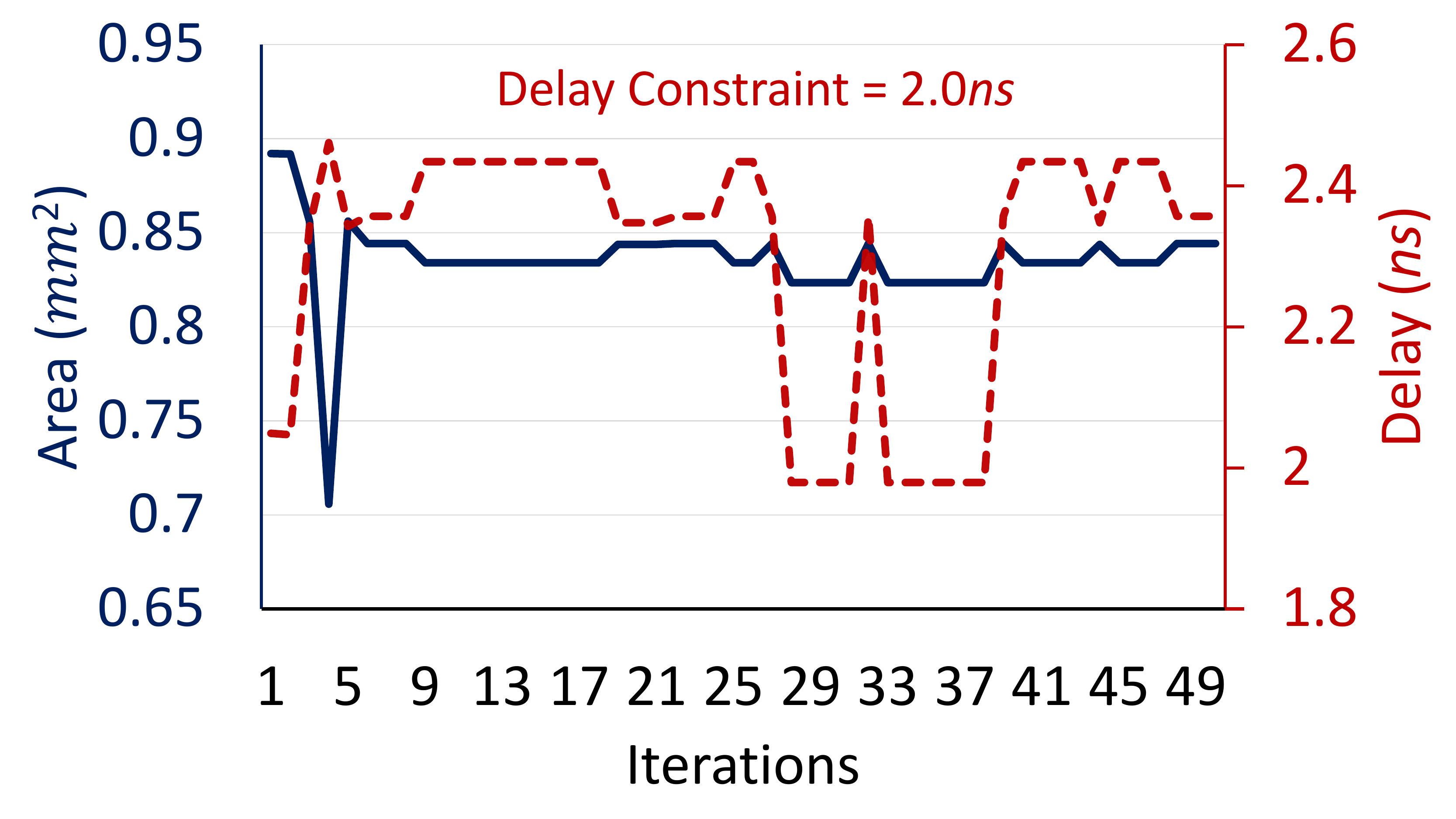}
        \caption{Adder}%
    \end{subfigure}
    \begin{subfigure}[t]{0.33\textwidth}
        \includegraphics[scale=0.17]{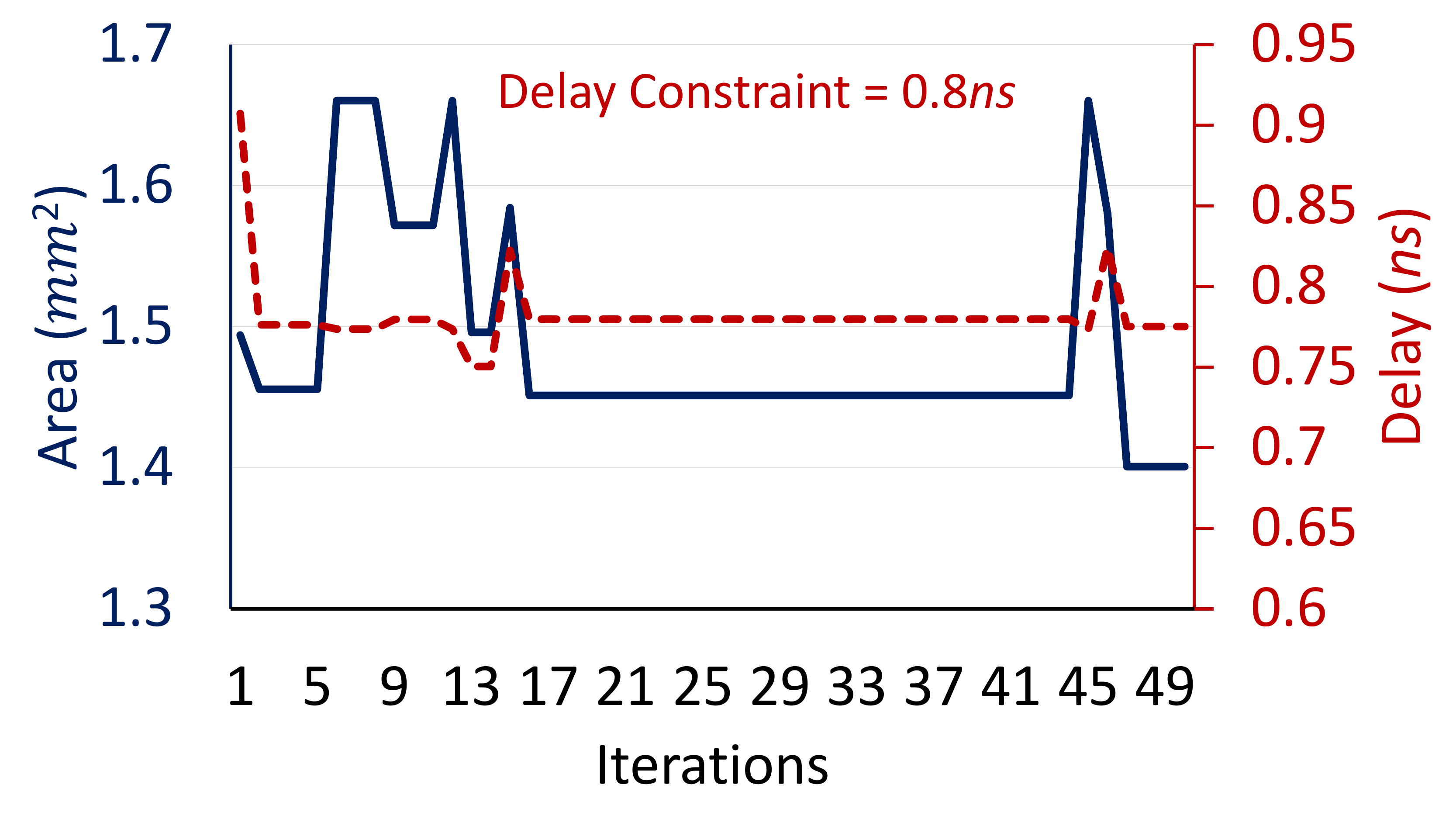}
        \caption{Barrel Shifter}
    \end{subfigure}
    \begin{subfigure}[t]{0.33\textwidth}
        \includegraphics[scale=0.17]{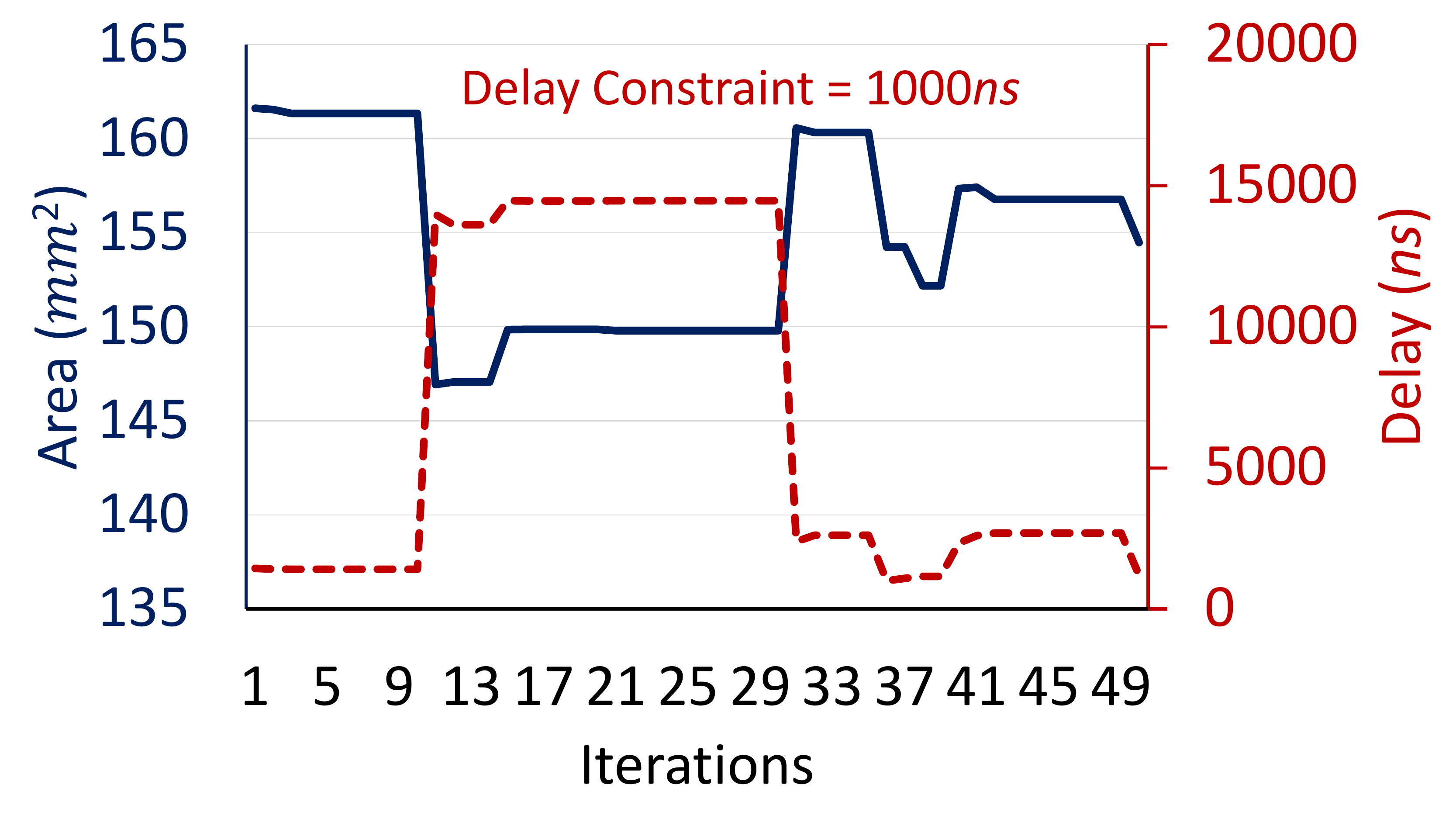}
        \caption{Hypotenuse}
    \end{subfigure}
    \caption{Traces of DRiLLS agent navigating the design space to find a design with a minimum area while meeting the delay constraint.}
    \label{fig:training-episodes}
\end{figure*}

%% file: 5results.tex
\begin{table*}[t!!]
  \small
  \centering
  \caption{Area-delay comparison of logic synthesis optimization results. A greedy algorithm optimizes for area. Expert-crafted scripts are derived from \cite{expertscripts}. EPFL best results for size are available at \cite{epfl}.}
  \label{table:summary}
  \setlength\tabcolsep{2.5pt}
  \def\arraystretch{1.35}%

\begin{tabular}{|r|r|r|r|r|r|r|r|r|r|r|r|r|r|r|r|}
\hline
\multirow{2}{*}{\textbf{Benchmark}} & \multicolumn{1}{c|}{\multirow{2}{*}{\textit{\textbf{\begin{tabular}[c]{@{}c@{}}Delay\\ Constr.\\ ($ns$)\end{tabular}}}}} & \multicolumn{2}{c|}{\textbf{Initial Design}} & \multicolumn{3}{c|}{\textbf{Greedy}} & \multicolumn{3}{c|}{\textbf{Expert-crafted \cite{expertscripts}}} & \multicolumn{3}{c|}{\textbf{EPFL Best Size \cite{epfl}}} & \multicolumn{3}{c|}{\textbf{DRiLLS}} \\ \cline{3-16} 
 & \multicolumn{1}{c|}{} & \multicolumn{1}{c|}{\textit{\textbf{\begin{tabular}[c]{@{}c@{}}Area\\ ($um^2$)\end{tabular}}}} & \multicolumn{1}{c|}{\textit{\textbf{\begin{tabular}[c]{@{}c@{}}Delay\\ ($ns$)\end{tabular}}}} & \multicolumn{1}{c|}{\textit{\textbf{\begin{tabular}[c]{@{}c@{}}Area\\ ($um^2$)\end{tabular}}}} & \multicolumn{1}{c|}{\textit{\textbf{\begin{tabular}[c]{@{}c@{}}Delay\\ ($ns$)\end{tabular}}}} & \multicolumn{1}{c|}{\textit{\textbf{\begin{tabular}[c]{@{}c@{}}Impr.\\ (\%)\end{tabular}}}} & \multicolumn{1}{c|}{\textit{\textbf{\begin{tabular}[c]{@{}c@{}}Area\\ ($um^2$)\end{tabular}}}} & \multicolumn{1}{c|}{\textit{\textbf{\begin{tabular}[c]{@{}c@{}}Delay\\ ($ns$)\end{tabular}}}} & \multicolumn{1}{c|}{\textit{\textbf{\begin{tabular}[c]{@{}c@{}}Impr.\\ (\%)\end{tabular}}}} & \multicolumn{1}{c|}{\textit{\textbf{\begin{tabular}[c]{@{}c@{}}Area\\ ($um^2$)\end{tabular}}}} & \multicolumn{1}{c|}{\textit{\textbf{\begin{tabular}[c]{@{}c@{}}Delay\\ ($ns$)\end{tabular}}}} & \multicolumn{1}{c|}{\textit{\textbf{\begin{tabular}[c]{@{}c@{}}Impr.\\ (\%)\end{tabular}}}} & \multicolumn{1}{c|}{\textit{\textbf{\begin{tabular}[c]{@{}c@{}}Area\\ ($um^2$)\end{tabular}}}} & \multicolumn{1}{c|}{\textit{\textbf{\begin{tabular}[c]{@{}c@{}}Delay\\ ($ns$)\end{tabular}}}} & \multicolumn{1}{c|}{\textit{\textbf{\begin{tabular}[c]{@{}c@{}}Impr. \\ (\%)\end{tabular}}}} \\ \hline
\textbf{Adder} & 2.00 & 867 & 2.02 & 1011 & 4.10 & -16\% & 1772 & 1.82 & -104\% & 1690 & 1.87 & -94\% & 823 & 1.97 & 5\% \\ \hline
\textbf{B. Shifter} & 0.80 & 2499 & 1.03 & 2935 & 0.66 & -17\% & 1534 & 0.77 & 38\% & 1040 & 0.77 & 58\% & 1400 & 0.77 & 43\% \\ \hline
\textbf{Divisor} & 75.00 & 12388 & 75.83 & 22439 & 79.14 & -81\% & 21167 & 65.05 & -70\% & 16031 & 74.91 & -29\% & 13441 & 67.61 & -8\% \\ \hline
\textbf{Hypotenuse} & 1000.00 & 176938 & 1774.32 & 236271 & 563.12 & -33\% & 210828 & 525.34 & -19\% & 169468 & 1503.88 & 4\% & 154227 & 995.95 & 12\% \\ \hline
\textbf{Log2} & 7.50 & 19633 & 7.63 & 30893 & 6.96 & -57\% & 18451 & 7.45 & 6\% & 23999 & 10.12 & -22\% & 17687 & 7.44 & 9\% \\ \hline
\textbf{Max} & 4.00 & 1427 & 4.48 & 3082 & 3.79 & -115\% & 1440 & 3.93 & -0.88\% & 1713 & 4.84 & -20\% & 1037 & 3.76 & 27\% \\ \hline
\textbf{Multiplier} & 4.00 & 19617 & 3.83 & 25219 & 4.38 & -28\% & 21094 & 3.70 & -7\% & 19940 & 5.27 & -1\% & 17797 & 3.96 & 9\% \\ \hline
\textbf{Sin} & 3.80 & 3893 & 3.65 & 5501 & 2.88 & -41\% & 4421 & 2.19 & -13\% & 4892 & 4.14 & -25\% & 3050 & 3.76 & 21\% \\ \hline
\textbf{Square-root} & 170.00 & 11719 & 329.46 & 19233 & 93.71 & -64\% & 16594 & 92.30 & -41\% & 9934 & 169.46 & 15\% & 9002 & 167.47 & 23\% \\ \hline
\textbf{Square} & 2.20 & 11157 & 2.27 & 19776 & 3.96 & -77\% & 16373 & 1.59 & -46\% & 16838 & 4.06 & -50\% & 12584 & 2.199 & -12\% \\ \hline
\multicolumn{2}{|l|}{\textbf{Avg. Area Imprv.}} & \multicolumn{2}{c|}{0.00\%} & \multicolumn{3}{c|}{-53.31\%} & \multicolumn{3}{c|}{-26.00\%} & \multicolumn{3}{c|}{-16.69\%} & \multicolumn{3}{c|}{13.19\%} \\ \hline
\multicolumn{2}{|l|}{\textbf{Constraint Met}} & \multicolumn{2}{c|}{2/10} & \multicolumn{3}{c|}{4/10} & \multicolumn{3}{c|}{9/10} & \multicolumn{3}{c|}{4/10} & \multicolumn{3}{c|}{10/10} \\ \hline
\end{tabular}

\end{table*}

We demonstrate the proposed methodology by utilizing the open-source synthesis framework ABC v1.01~\cite{mishchenko2007abc}. We implement DRiLLS in Python v3.5.2 and utilize TensorFlow r1.12~\cite{tensorflow} to train the A2C agent neural networks.
All experiments are synthesized using ASAP7, a 7 nm standard cell library in typical processing corner.
We evaluate our framework on EPFL arithmetic benchmarks~\cite{epfl}, exhibiting wide ranges of circuit characteristics. The characteristics of the evaluated benchmarks (e.g. I/Os, number of nodes, edges and levels) can be found in~\cite{epfl}. Experimental parameters were setup as:

\begin{itemize}
    \item \textbf{Episodes ($N$):} 50, \textbf{Iterations ($k$):} 50
    \item \textbf{Networks Size: } \textit{Actor: } 2 fully connected layers, 20 hidden units each. \textit{Critic:} one hidden layer with 10 units.
    \item \textbf{Weight initialization:} Xavier initialization ~\cite{xavier}
    \item \textbf{Optimizer:} Adam ~\cite{adam}, \textbf{Learning Rate: ($\alpha$)}: 0.01
    \item \textbf{Discount rate ($\gamma$):} 0.99
\end{itemize}

\noindent A small number of layers is used as we observe that deeper neural networks exhibit a random behavior and do not train well in this framework. This is attributed to the nature of the small number of features and transformations used. The experimental results are obtained using a machine with Intel Xeon 2x14cores@2.4 GHz, 128GB RAM, and 1x500GB SSD; running Ubuntu 16.04 LTS. Next, we present our results.

\subsection{Design Space Exploration}
Figure \ref{fig:training-episodes} shows traces of the agent searching for an optimized design that minimizes area, and meets the delay constraint. We plot one episode that finds the global minimum for a number of representative benchmarks. Generally, Figure \ref{fig:training-episodes} shows the attempts of the agent to balance between reducing the design area and meeting the delay constraint. For example, we observe the various trials of the agent to execute a transformation that reduces the delay to meet the constraint, but increases the design area such as iteration 30 in Log2 and iteration 26 in Max. Occasionally, exploration saturates as we can notice near-straight lines in some iterations. This shows the ability of the actor-critic networks to guide the exploration, while occasionally exploring other transformations that might open new search paths.

\subsection{Comparison to Other Techniques}
We compare the agent's performance against {\it EPFL best results}, {\it expert-crafted scripts}, and a {\it greedy heuristic algorithm}:

\begin{enumerate}
    \item {\it EPFL best results}: best results are provided for size and depth. We compare against best results for size, since it is more relevant to the agent's nature of optimizing for area when mapping to a standard cell library.
    \item {\it Expert-crafted scripts}: we maintain a record of expert-crafted synthesis optimizations derived from \cite{expertscripts}.
    \item {\it Greedy heuristics algorithm}: we developed a baseline comparison that takes an initial input design and spawns parallel threads to perform each of the given AIG transformations on the design. Afterwards, each thread performs the mapping step using the delay constraint. The algorithm then evaluates the mapped designs from all threads, and keeps the one with the minimum area for the next iteration. After that, the process is repeated until two iterations yield the same area.
\end{enumerate}

Table \ref{table:summary} gives the results of the mentioned comparisons. The area and delay for the initial design are obtained by loading the non-optimized designs in ABC and mapping them to ASAP7 without performing any transformation on the AIG. The delay is reported using the built-in timer in ABC (using \textit{stime} command). We use the initial run to select a delay constraint value that challenges all the methods studied in this work. We make the following observations:

\begin{itemize}
    \item The greedy algorithm has a single optimization target (area). Although the delay constraint was met in 4 designs, it is attributed to the best-effort mapping step that considers the delay constraint. The increase in the area occurs in the first iteration that tries to meet the delay constraint while mapping. Since the algorithm meets the stop criteria in the first few iterations, it fails to reduce the area subject to a delay constraint. Results show the smallest average area improvement.
    \item Although expert-crafted synthesis scripts have not improved the designs' areas, they produced optimized designs that meet the delay constraint in 9 out of 10 designs. This comes at no surprise as the techniques used strive to meet the delay constraint; therefore, accepting near-optimal area results~\cite{expertscripts}.
    \item EPFL best results have shown decent improvements in 3 designs, meeting the delay constraint in 4 of them. Although we benchmarked on the best results in terms of size, not depth, it is reasonable that their optimization techniques have not been designed for standard cell library mapping.
    \item DRiLLS agent meets the delay constraint in all designs while simultaneously improving the design area by an average of $13.19\%$. In the two designs that DRiLLS increased their area, it in fact met the delay constraint which the un-optimized design did not meet. This proves that the reward function defined before is an effective one for training the agent. Moreover, DRiLLS outperforms EPFL best result in all designs except {\it Barrel shifter}.
\end{itemize}

\begin{figure*}[t!]
     \begin{subfigure}[t]{0.33\textwidth}
        \includegraphics[scale=0.18]{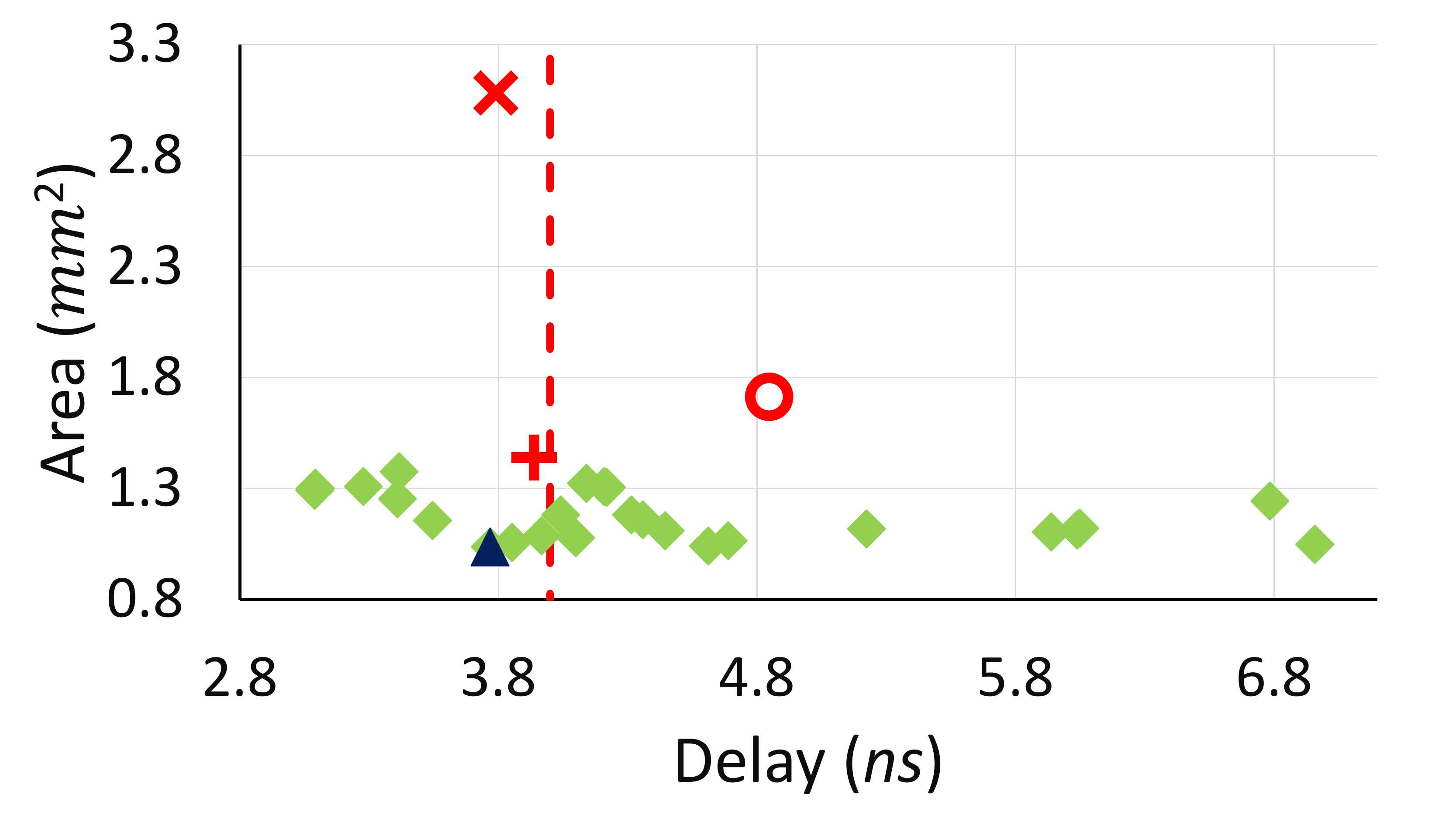}
        \caption{Max}%
    \end{subfigure}
    \begin{subfigure}[t]{0.33\textwidth}
        \includegraphics[scale=0.18]{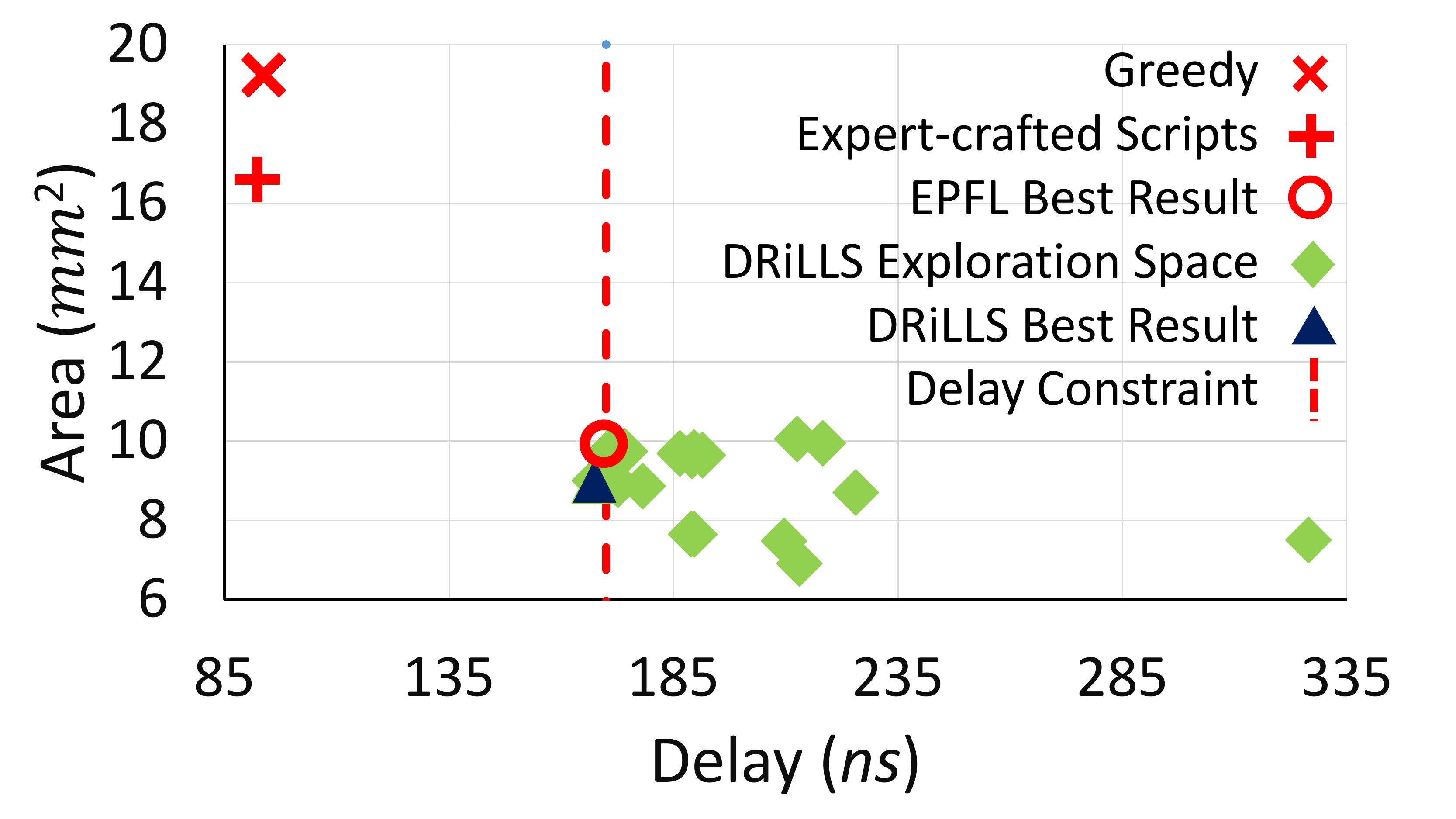}
        \caption{Square-root}
    \end{subfigure}%
    \begin{subfigure}[t]{0.33\textwidth}
        \includegraphics[scale=0.18]{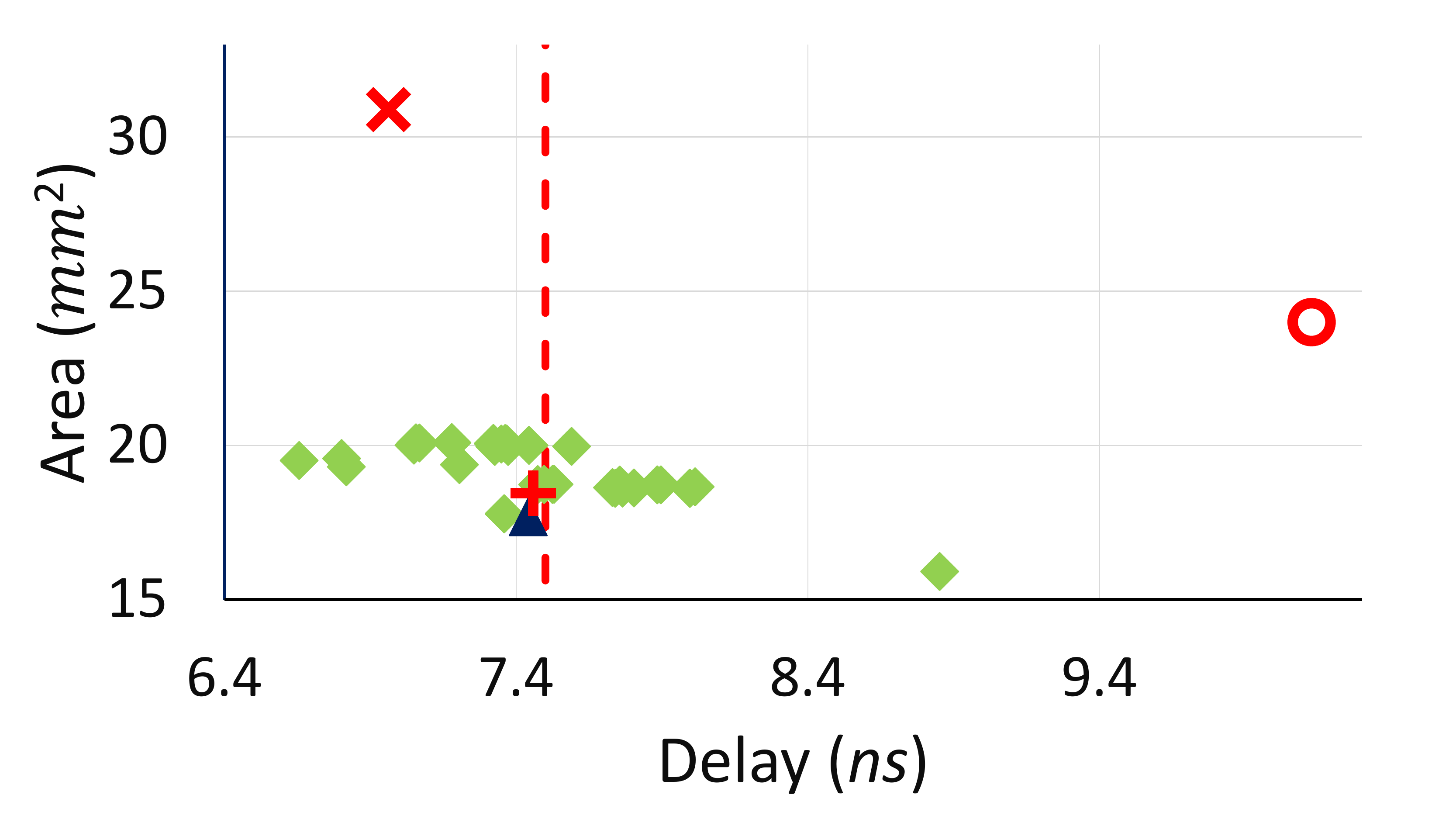}
        \caption{Log2}
    \end{subfigure}
    \begin{subfigure}[t]{0.33\textwidth}
        \includegraphics[scale=0.18]{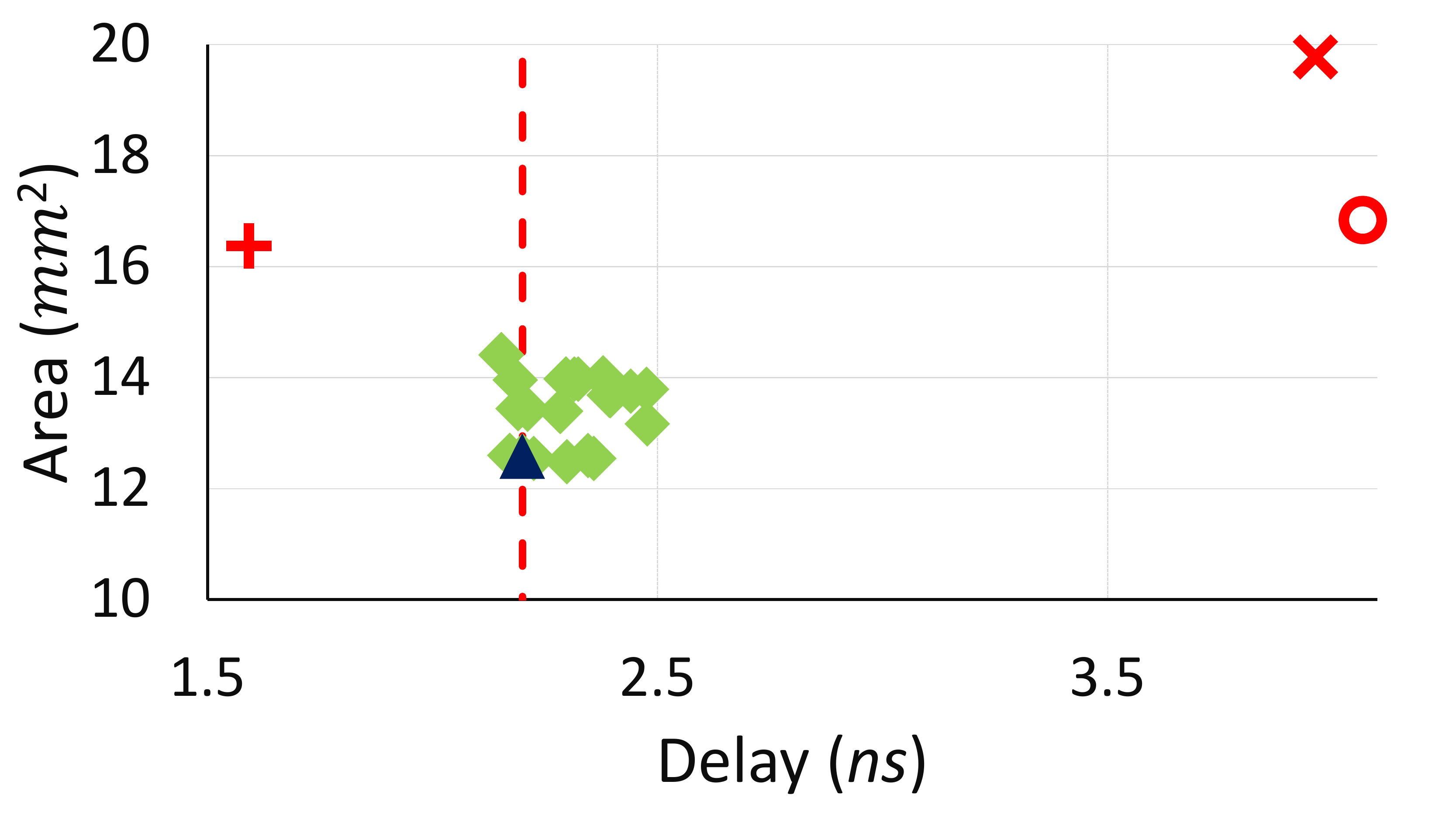}
        \caption{Sin}
    \end{subfigure}%
    \begin{subfigure}[t]{0.33\textwidth}
        \includegraphics[scale=0.18]{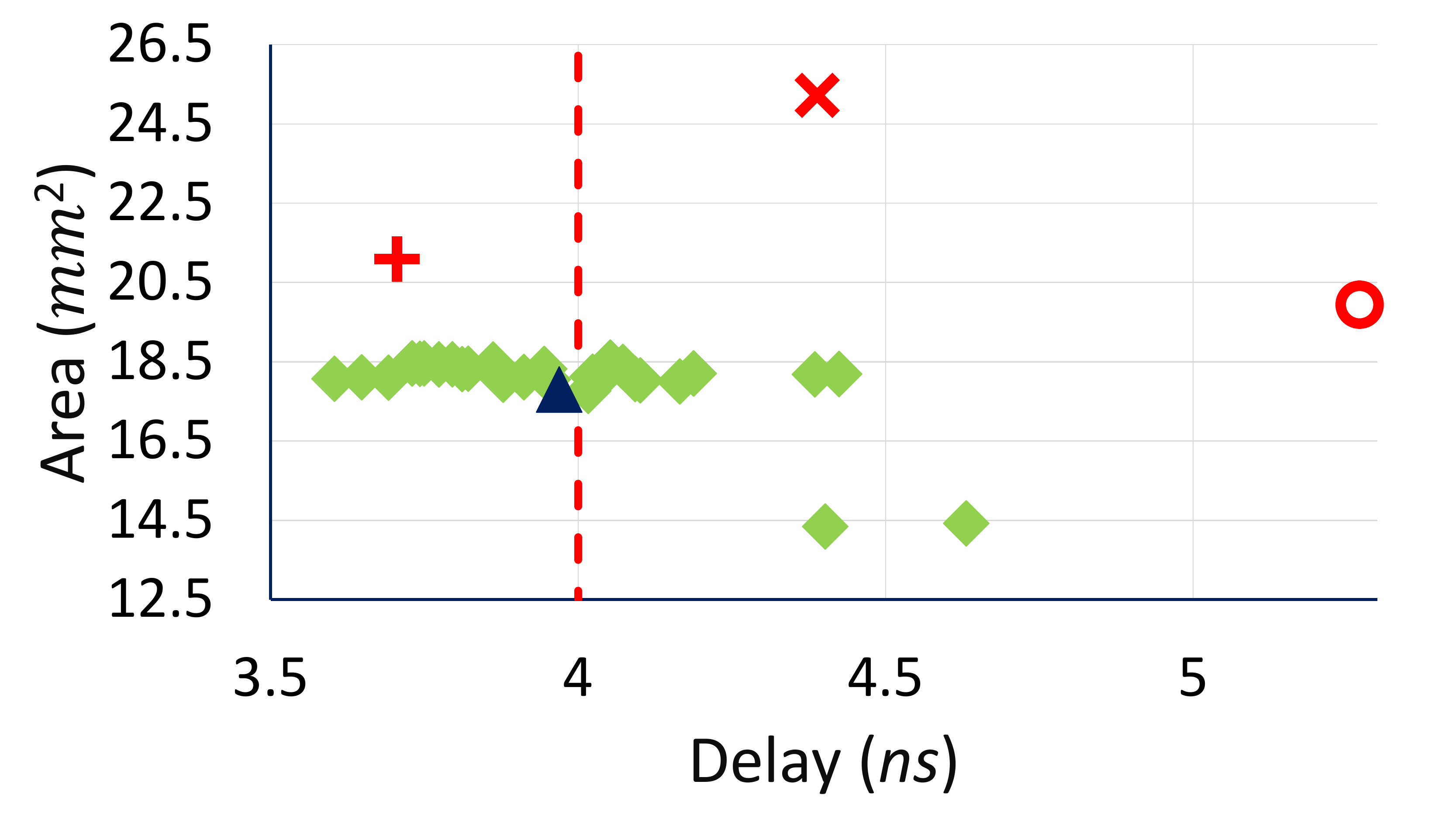}
        \caption{Multipler}%
    \end{subfigure}
    \begin{subfigure}[t]{0.33\textwidth}
        \includegraphics[scale=0.18]{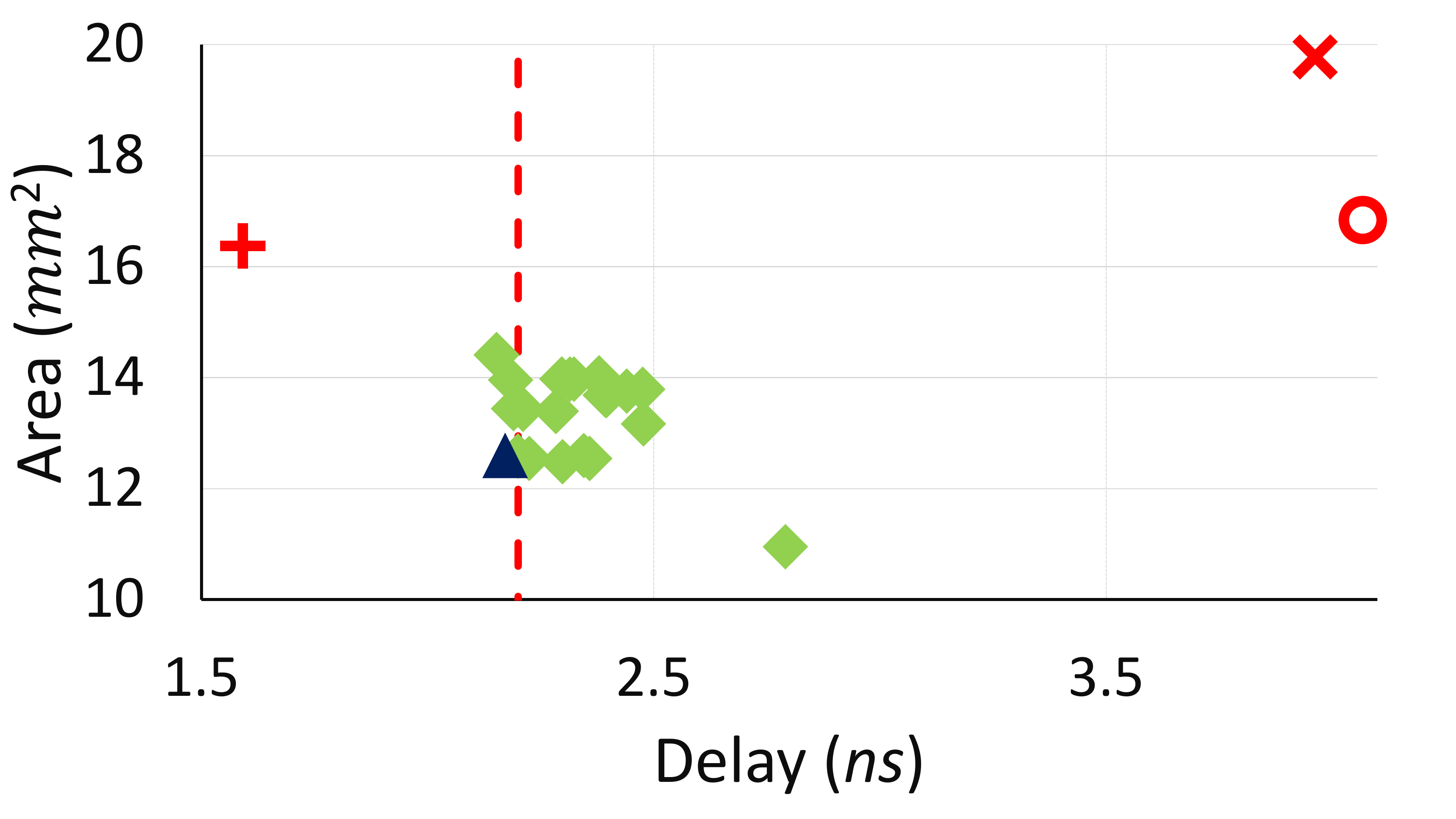}
        \caption{Square}
    \end{subfigure}
    \caption{Design area vs. delay trade-offs. The vertical dotted red line shows the delay constraint. For each benchmark, DRiLLS exploration space is indicated in green diamonds. A highlighted triangle represents the best optimized design that meets the delay constraint. Other methods are shown in red color with a cross mark, a plus mark and a circle for greedy, expert-crafted and EPFL result respectively.}
    \label{fig:scatter}
\end{figure*}

In the interest of space, Figure~\ref{fig:scatter} elaborates on Table \ref{table:summary} by plotting the area-delay trade-offs offered by DRiLLS against the greedy algorithm, the expert-crafted synthesis scripts and the EPFL best results on six of the benchmarks. We define the \textbf{exploration run time} as the total run time of the agent, including interacting with the {\it Logic Synthesis Environment}, extracting AIG characteristics, and optimizing the parameters of the agent networks. The smallest design (Adder) is explored in \textit{3.25mins}, while the largest (Hypotenuse) is explored in \textit{25.46mins}. The average exploration time is \textit{12.76mins} per episode. It is important to note that a trained model on one circuit design can be used (reloaded) into a new exploration on new circuits requiring no retraining.

%% file: 6conclusions.tex
The goal of developing DRiLLS is to offer an autonomous framework that is able to explore the optimization space of a given circuit design, and produce a high Quality of Result (QoR) with no human in-loop. The intuition behind modeling this problem into a reinforcement learning context is to provide the machine with a methodology to try and error, similar to how human experts gain their experience optimizing designs. 

In this work, we have presented a methodology based on reinforcement learning that enables autonomous and efficient exploration of the logic synthesis design space. Our proposed methodology maps the complex search space to a ``game'' where an advantage actor critic (A2C) agent learns to maximize its reward (reduce area subject to a delay constraint) by iteratively choosing primitive transformations with the highest expected reward. We have formulated an AIG state representation that has proved to effectively represent the feature set of a design state. In addition, we have introduced a novel multi-objective reward function that guides the exploration process of the agent. It allows the agent to find a minimum design area subject to delay constraint. Evaluating ten representative benchmarks, our proposed methodology manifests results that outperform existing methods. 

DRiLLS proves that Reinforcement Learning can be used in combinatorial optimization of hardware circuit designs. It has a broad potential to be applied on related physical synthesis tasks, eliminating the need for human experts. The framework is open-source under a permissive license (BSD-3) and is available publicly on GitHub\footnote{https://github.com/scale-lab/DRiLLS}.